\documentclass[letterpaper, 10 pt, conference]{ieeeconf}  

\IEEEoverridecommandlockouts                              

\UseRawInputEncoding

\usepackage{graphics} 
\usepackage{epsfig} 
\usepackage{mathptmx} 
\usepackage{times} 
\usepackage{amsmath} 
\usepackage{amssymb}  
\usepackage{bm}
\usepackage{subfigure} 
\usepackage{flushend} 
\usepackage{multirow}
\usepackage{booktabs}

\title{\LARGE \bf
Stereo Plane SLAM Based on Intersecting Lines
}

\author{Xiaoyu Zhang, Wei Wang, Xianyu Qi and Ziwei Liao
\thanks{This work was supported by the National Key Research and Development Program of China under grant number 2020YFB1313600.}
\thanks{The authors are with the Robotics Institute, Beihang University, Beijing, China. (Email: zhang\_xy, wangweilab, qixianyu, liaoziwei @buaa.edu.cn)}%
}

\begin{document}
\maketitle
\thispagestyle{empty}
\pagestyle{empty}

\begin{abstract}

Plane features can be used to reduce drift errors in SLAM systems, especially in indoor environments. It is easy and efficient to extract planes from a dense point cloud, which is commonly generated from a RGB-D camera or a 3D lidar. But when using a stereo camera, it is hard to compute dense point clouds accurately or efficiently. In this paper, we propose a novel method to compute plane parameters using intersecting lines, which are extracted from stereo images. Plane features are commonly extracted from the surface of man-made objects or structures, which have regular shapes and straight edge lines. In three dimensions, two intersecting lines determine a unique plane. Therefore, we extract line segments from both left and right images of a stereo camera. By stereo matching, we compute lines' endpoints and direction vectors, and then a plane from two intersecting lines is calculated. We discard inaccurate plane features in the frame tracking. Adding such plane features in the stereo SLAM system reduces drift errors and refines the performance. Finally, we build a global map consisting of both points and planes, which can reflect real scene structures. We test our proposed system on public datasets and demonstrate its accurate estimation results, compared with state-of-the-art SLAM systems. To benefit the research of plane-based SLAM, we release our codes at https://github.com/fishmarch/Stereo-Plane-SLAM.

\end{abstract}

\section{INTRODUCTION}
Simultaneous localization and mapping (SLAM) is a fundamental problem for various applications, including robots, driverless cars, and augmented reality (AR). Thanks to the increasingly powerful capability of graphics processing, cameras have been widely used and visual SLAM has developed rapidly in the past decade \cite{Cadena2016Past}.

For visual SLAM, point is the most commonly used feature to track camera poses and build environmental models. Some existing point-based SLAM systems \cite{Mur-Artal2017,Engel2018} have achieved accurate and robust estimation results. Except for points, lines \cite{Gomez-Ojeda2019} and planes \cite{kaess2015simultaneous} are also explored to improve SLAM performance in recent years. It is proven that lines and planes are helpful to build more robust and accurate SLAM systems, especially in indoor environments. Line features are commonly extracted using the Line Segment Detector \cite{Gioi2010LSD} from images. Line segments can be matched by LBD descriptor \cite{Zhang2013An} in frame tracking, but they suffer from the problem of occluding and endpoint variance. Besides, the parameterization of line segments \cite{hartley2003multiple} is complicated.

\begin{figure}[thpb]
	\centering
	\includegraphics[width = 8.5cm]{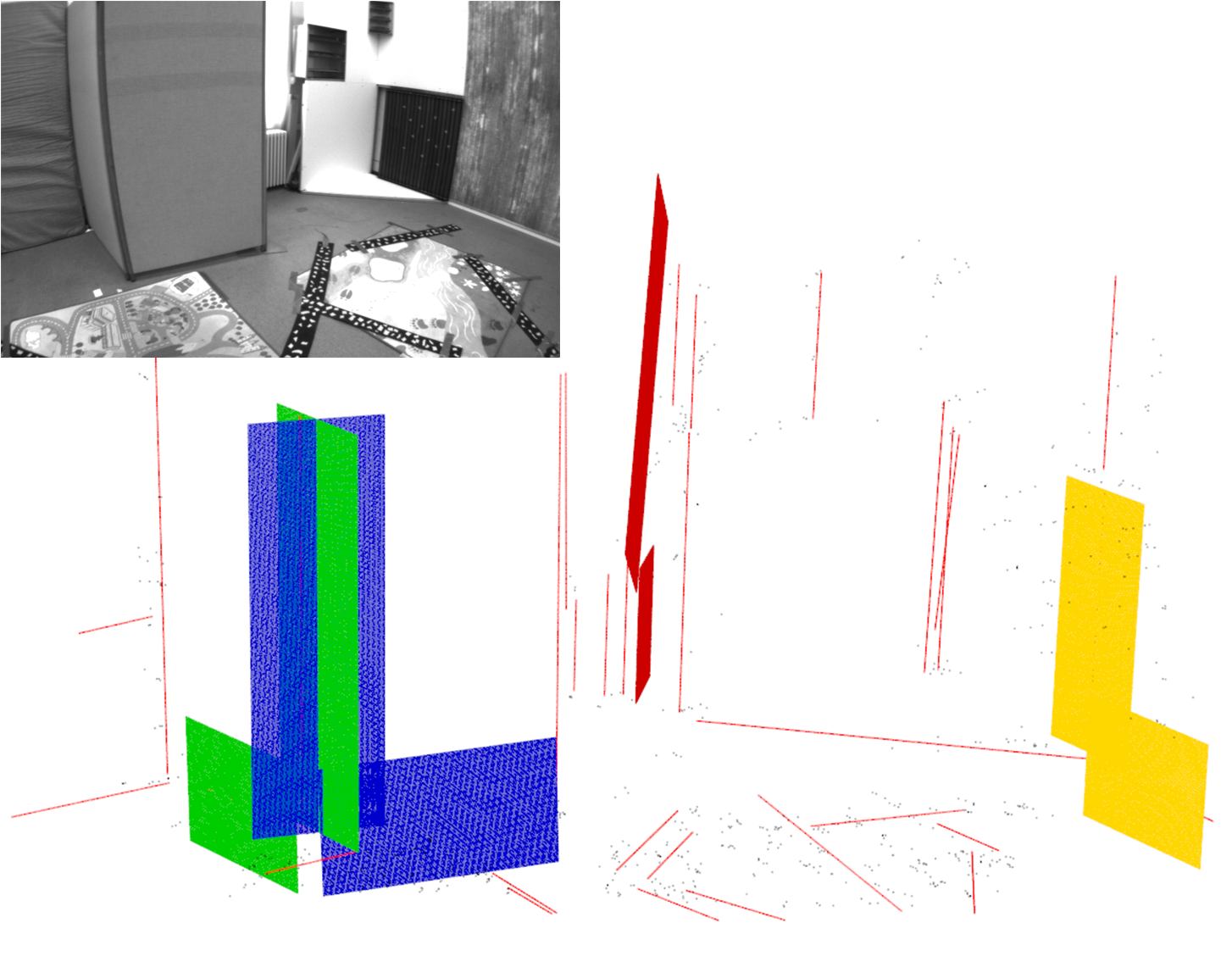}
	\caption{The plane features computed from intersecting lines. The planes are drawn in different colors. The black points are point features. The line segments are drawn in red for illustration, but they are not included in our SLAM system. }
	\label{3dplane}
\end{figure}

Compared with line features, planes are commonly more robust features in SLAM because of their simple and accurate data association. Planes can even be matched in frames of a large distance, which helps reduce drift errors. Besides, planes are helpful to reflect real scene structures. Plane features are commonly extracted from a dense point cloud \cite{trevor2013efficient}, generated by a RGB-D camera or a 3D lidar. In stereo images, however, it is not easy to get a dense point cloud accurately or efficiently. The depth of a point can be estimated by matching corresponding pixels in stereo images. But matching all pixels becomes a tough task. Some traditional stereo matching methods \cite{4359315} utilize low-level features of image patches to match pixels. They run fast but suffer from low quality. Recently, stereo matching algorithms based on deep learning have achieved remarkable performance \cite{Kun2020Review}. But these methods are slow and need high-cost GPU.

This paper proposes a novel method to compute plane features from stereo images. Plane features are commonly extracted from the surface of man-made objects and structures. These planes commonly have regular shapes and straight edge lines. In three dimensions, two intersecting lines determine a unique plane. Therefore, it is reasonable and possible to compute plane features from lines. 3D lines can be computed from stereo images by stereo matching \cite{Gomez-Ojeda2019}. An example of computing plane features is shown in Fig. \ref{3dplane}, the original scene is also shown at the top-left corner. We also provide a video at https://youtu.be/3VWF-JJU9T8. 

In summary, our contributions are as follows:
\begin{itemize}
	\item A novel method to compute plane features from stereo images based on intersecting lines.
	\item A stereo SLAM system using extracted points and computed planes.
	\item Evaluated on public datasets, our system gets accurate estimation results and achieves state-of-the-art performance.
\end{itemize}

In the following, we first introduce the related work in Sec. \ref{relatework}, then explain the method to compute plane features in Sec. \ref{plane sec}, followed by the introduction of the whole system in Sec. \ref{systemSec}. In the end, we show our experimental results in Sec. \ref{evaluationSec}.

\section{RELATED WORK}
\label{relatework}
SLAM is well studied and different methods have been proposed in recent years. Many SLAM systems are commonly based on point features and build a global map consisting of 3D points. ORB-SLAM \cite{Mur-Artal2017} tracks ORB features and uses re-projection error to estimate camera poses. In contrast, direct methods \cite{LSDSLAM} use photometric error to track camera poses. Point features can be used to build a sparse map \cite{DSO}, semi-dense map \cite{SVO} and even dense map \cite{DTAM}. In these methods, it is easy to extract and compute point features, and they work well in most scenes.

In recent years, researchers also utilize line and plane features to refine the performance of SLAM systems, especially for indoor environments. In a SLAM system, line and plane features work like a filter to reduce the measurement errors, and perform well even in some low-texture scenes. Pumarola et al. \cite{pumarola2017pl} extract both point and line features from monocular images and improve SLAM performance for low-texture scenes. Zhang et al. \cite{Zhang2015} propose a 3D line-based SLAM system using a stereo camera and exhibit its better reconstruction performance. PL-SLAM \cite{Gomez-Ojeda2019} is built based on ORB-SLAM, and also leverages both points and line segments. Qian et al. \cite{Qian2020} extend the work of PL-SLAM using Bags of Point and Line Word and release their codes.

Compared with line features, planes are more accurate and robust landmarks. Taguchi et al. \cite{taguchi2013point}  present a framework for registration combining points and planes. CPA-SLAM \cite{CPASLAM} proposes a novel formulation to track camera poses using global planes in the expectation-maximization (EM) framework. Kaess et al. \cite{kaess2015simultaneous} introduce a minimal representation for infinite planes which is suitable for the least-squares estimation without encountering singularities. \cite{Zhang2019} explores both plane and plane edges to increase plane observations and constraints. These plane-based SLAM systems also achieve more robust and accurate estimation results compared with point-based methods, especially in low-texture scenes. All of these systems are based on RGB-D cameras, from which planes are easily extracted. Our work utilizes some ideas from line-based SLAM, and extends plane-based SLAM to using stereo cameras.

\begin{figure}[t]
	\centering	
	\subfigure[Matching of line segments]{
		\centering
		\includegraphics[width = 8.5cm]{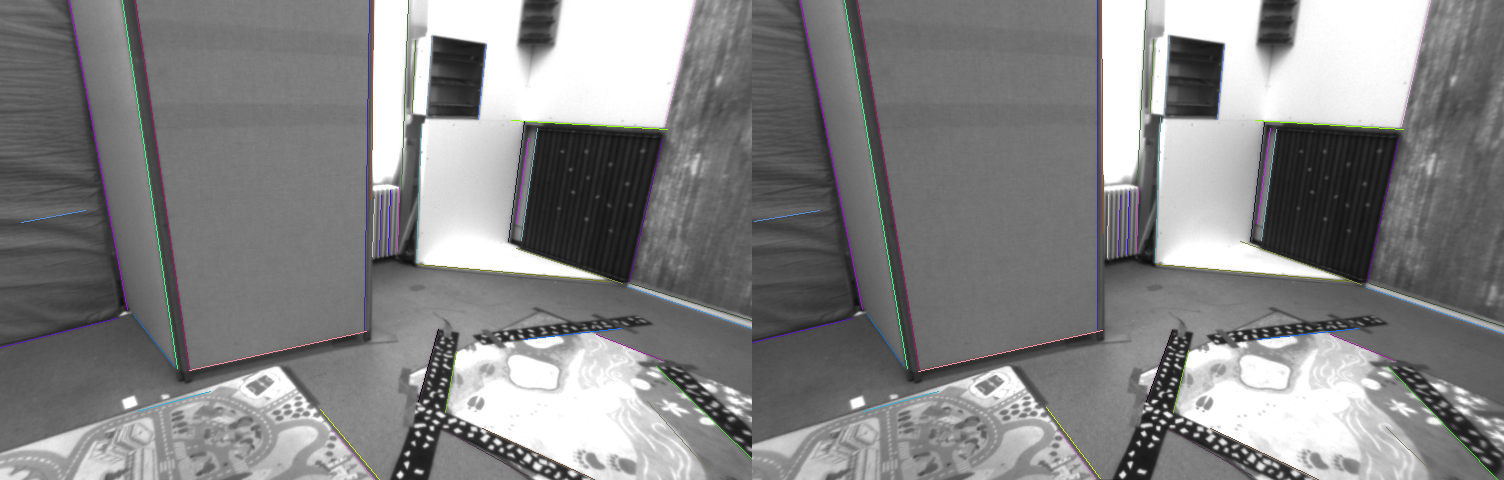}
		\label{linesegmentmatch}
	}
	\subfigure[Matching of endpoints]{
		\centering
		\includegraphics[width = 8.5cm]{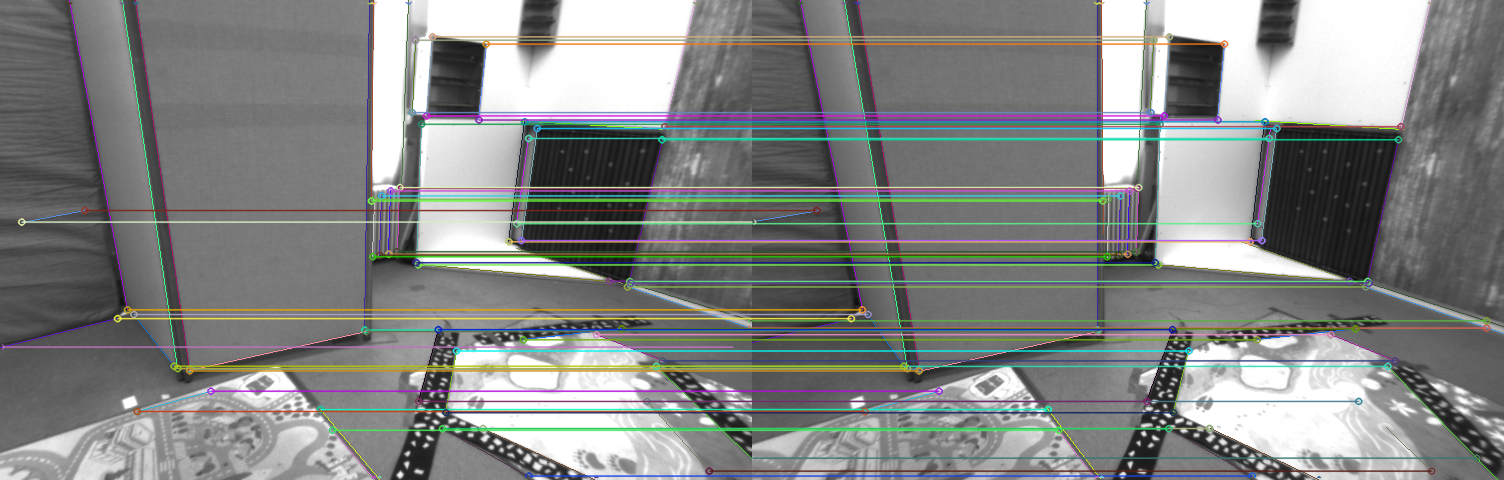}
		\label{segmentmatch}
	}
	\caption{Matching of line segments and endpoints. }
	\label{linematch}
\end{figure}

\section{PLANE FEATURES FROM INTERSECTING LINES}
\label{plane sec}
This section is to introduce the method of computing plane features. We first extract line segments from both left and right images of a stereo camera. By matching line segments and their endpoints, we compute 3D positions of endpoints and lines' direction vectors. Then we check their position to find intersecting lines. Finally, we compute plane parameters.

\subsection{Notations}
We denote a plane feature as $ {\pi} = \left(\boldsymbol{n}^{\top}, d \right)^{\top} $, where $ \boldsymbol{n}=\left(n_x,n_y,n_z\right)^{\top} $ is the unit normal vector representing the plane's orientation and $ d $ is the distance of the plane from the origin. We use the commonly used form $\boldsymbol{T}_{cw} \in {S E}(3)$ to represent a camera pose and $\boldsymbol{p}=\left(x, y, z, 1\right)^{\top}$ to represent a 3D point. Therefore, $\boldsymbol{T}_{cw} \boldsymbol{p}_{w} $ transforms a point from the world to the camera coordinate system and $ \boldsymbol{T}_{cw}^{-\top}{\pi}_{w} $ transforms a plane from the world to the camera coordinate system.

For lines, we only record their endpoints $(\boldsymbol{p}_s, \boldsymbol{p}_e)$ and unit direction vectors $\boldsymbol{n}_l$, which are enough to compute plane features.

\subsection{Line Detection and Computation}

A frame from a stereo camera consists of a left image $I_l$ and a right image $I_r$ . We use the Line Segment Detector \cite{Gioi2010LSD} to extract line segments from both $I_l$ and $I_r$, and match them by the LBD descriptor \cite{Zhang2013An}. The line matching is accurate and robust enough in one stereo frame. As shown in Fig. \ref{linesegmentmatch}, the line segments are drawn in different colors, and the matched line segments are the same color in $I_l$ and $I_r$.

For every matched line segment in the left image $I_l$, we find corresponding points of its endpoints in the right image $I_r$ based on the fact that their row positions remain the same in $I_l$ and $I_r$ when using a parallel stereo camera. As shown in Fig. \ref{segmentmatch}, matched endpoints are connected by transverse lines. 

From the stereo matching of endpoints, we compute their 3D positions $\boldsymbol{p}$ based on disparities $\Delta u$. A line's direction vector $\boldsymbol{n}_l$ is also defined by its two endpoints.

\subsection{Plane Computation}

Before computing plane features, we need to check the relationship of lines. In three dimensions, two intersecting or parallel lines are on the same plane. For parallel lines, however, it is difficult to judge whether they are extracted from a real plane, and the planes computed from them are prone to bring large errors. Therefore, we only compute planes from intersecting lines. 

To find intersecting lines quickly, we find the lines meeting the following conditions:
\begin{itemize}
	\item The angle between two lines is larger than the threshold ($10^{\circ}$ in our experiments) 
	\item The distance between their central points is smaller than the length of the line.
	\item The four endpoints of these two lines lie on the same plane.
\end{itemize}

The central points $\boldsymbol{p}_c$ in the second condition are computed from lines' endpoints $\boldsymbol{p}_s$ and $\boldsymbol{p}_e$. From the first two conditions, we actually find unparallel lines that are close with each other. We compute the plane's normal vector by the cross product of the lines' direction vectors, 
$$
\boldsymbol{n}_{\pi} = \boldsymbol{n}_{li} \times \boldsymbol{n}_{lj} \eqno{(1)}
$$

Using the plane's normal $\boldsymbol{n}_{\pi}$ and four endpoints $\boldsymbol{p}_k$ $(k=1,2,3,4)$, we then compute four different plane coefficients $d_k$, 
$$
d_k = -\boldsymbol{n}_{\pi}\cdot\left(p_{kx},p_{ky},p_{kz}\right)^{\top} \eqno{(2)}
$$

The distance among them is:
$$
 D = Max\left(d_k\right) - Min\left(d_k\right) \eqno{(3)}
$$

If $D$ is smaller than the threshold (5 cm in our experiments), these two lines meet the third condition and the plane coefficients $ {\pi} = \left(\boldsymbol{n}_{\pi}^{\top}, \bar{d_k} \right)^{\top} $ are also computed, $\bar{d_k}$ here is the arithmetic average of $d_k$. Sometimes the computed planes may not be real planes, such as the plane from the lines of a doorframe. But such planes are also stable enough and provide accurate constraints, and therefore we treat them as real planes. 

Under these conditions, we compute as many planes as possible at first. We will validate the computed planes in the frame tracking and label inaccurate planes as invalid, which is described in Sec. \ref{validation}.

\begin{figure}[t]
	\centering
	\includegraphics[width = 8.5cm]{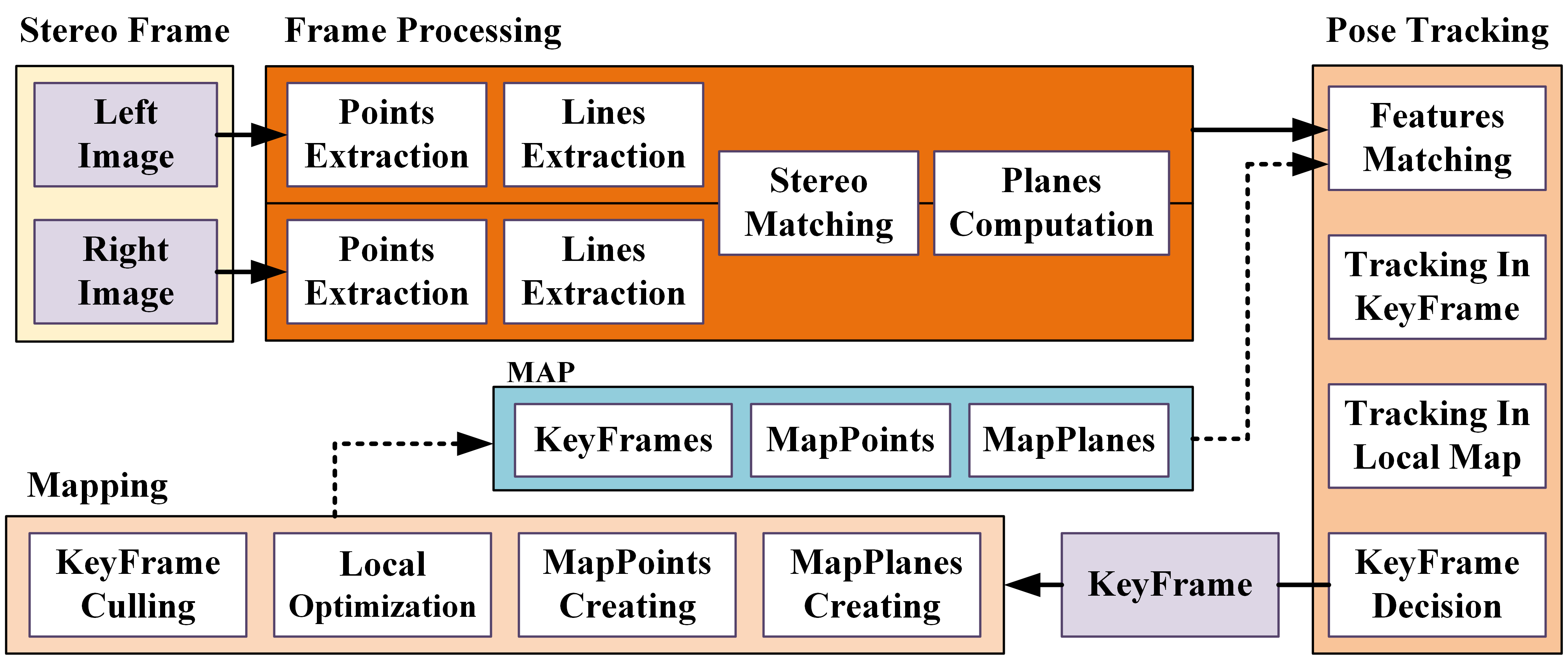}
	\caption{The pipeline of our proposed SLAM system.}
	\label{system}
\end{figure}

The features extracted and computed from the images of Fig. \ref{linematch} is shown in Fig. \ref{3dplane}. The black points are feature points extracted from the images. The red lines are the line segments extracted from the images, whose 3D positions are computed from the matched endpoints. Notice that we do not use these line segments in our SLAM system, and we draw them in the figure just to show the process of computing plane features. We draw the plane features by expanding corresponding intersecting lines and they are drawn in different colors. The red plane in the middle of the figure seems not correct because of errors from line segments, and we label this plane as invalid in the validation. But other planes are correct and are used as valid landmarks in our SLAM system.
 
\section{SLAM SYSTEM BASED ON COMPUTED PLANES}
\label{systemSec}
Points and planes are both used as landmarks and optimized in our SLAM system. We build our system based on the publicly available ORB-SLAM stereo version \cite{Mur-Artal2017}, which includes feature tracking and bundle adjustment optimization \cite{Triggs1999Bundle}. 

\subsection{System Overview}
The pipeline of our proposed SLAM system is illustrated in Fig. \ref{system}. It can be divided to three parts, frame processing, pose tracking and mapping. We do not add loop closure now, which will be a part of our future work.

In stereo frame processing, we extract feature points and line segments from both left and right images, and match these features based on descriptors. Then we can compute plane features using the method described in Sec. \ref{plane sec}.

In pose tracking, every camera pose is estimated based on matched valid features. The camera pose is firstly estimated from the last keyframe and then optimized in the local map.

In mapping, points and planes are constructed and saved in the map. To achieve more accurate estimation, a local map optimization is performed.

\subsection{Optimization Formulation}
SLAM is commonly formulated as a nonlinear least squares optimization problem \cite{Cadena2016Past}, and the bundle adjustment (BA) is commonly used for point features \cite{Mur-Artal2017}. Like points, we also design the optimization formulation for plane features. In our SLAM system, we denote a set of camera poses, point features and plane features as $C = \{{c}_i\}$, $P = \{{p}_j\}$, $\Pi = \{{\pi}_k\}$ respectively, then the optimization problem can be formulated as:
$$
\begin{aligned}
C^{*}, P^{*}, \Pi^{*}=\underset{\{C, P, \Pi\}}{\arg \min } &\sum_{{c}_{i}, {p}_{j}} \left\|{e}\left({c}_{i}, {p}_{j}\right)\right\|_{\Sigma_{i j}}^{2}+\\
&\sum_{{c}_{i}, {\pi}_{k}} \left\|{e}\left({c}_{i}, {\pi}_{k}\right)\right\|_{\Sigma_{i k}}^{2}  
\end{aligned}
\eqno{(4)}
$$
${e}\left({c}, {p}\right)$, ${e}\left({c}, {\pi}\right)$ represent the camera-point and camera-plane measurement errors respectively. $\|\boldsymbol{x}\|_{{\Sigma}}^2$ is the Mahalanobis distance, which equals $\boldsymbol{x}^{\top} {\Sigma}^{-1} \boldsymbol{x}$. $ {\Sigma} $ is the corresponding covariance matrix, it is set based on the measurement uncertainty (such as $2^{\circ}$ for plane angles in our experiments).

The optimization problem can be solved using Levenberg-Marquardt or Gauss-Newton implemented in g2o \cite{Rainer}.
\subsection{Measurement Error}
\subsubsection{Camera-Point Error}
We use the standard re-projection error for the camera-point measurement in our system:

$$
\boldsymbol{e}_{cp}\left(\boldsymbol{T}_{cw}, \boldsymbol{p}_{w} \right)=\boldsymbol{u}_c - \rho\left(\boldsymbol{T}_{cw}\boldsymbol{p}_{w}\right)
\eqno(5)
$$
Here $\boldsymbol{T}_{cw}$ is the camera pose, $\boldsymbol{p}_{w}$ is the point in the world coordinate system, $\boldsymbol{u}_c$ is the corresponding image pixel, and $\rho$ is the camera model to project 3D points onto the images. In optimization, the camera pose $\boldsymbol{T}_{cw}$ is mapped to Lie algebra $\boldsymbol{\xi} \in \mathfrak{s e}(3)$ to avoid extra constraints \cite{barfoot2017state}. The computation of the corresponding Jacobian matrix can also be found in \cite{barfoot2017state}. 
\subsubsection{Camera-Plane Error}
As a 3D plane has only three degrees of freedom, $ \pi = \left(\boldsymbol{n}^{\top}, d \right)^{\top} $ is over-parameterization. Therefore, it requires extra constraints to ensure the unit length of planes' normal vectors, adding additional computation in optimization. To overcome this problem, we follow the work in \cite{CPASLAM} using the minimal parameterization of planes ${\tau}=(\phi, \psi, d)^{\top}$ in optimization, where $ \phi $ and $ \psi $ are the azimuth and elevation angles of the normal vectors respectively:
$$
\tau=q(\pi)=\left(\phi=\arctan \frac{n_{y}}{n_{x}}, \psi=\arcsin n_{z}, d\right)^{\top}
\eqno(6)
$$

Then we define the measurement error using the minimal parameterization:
$$
\boldsymbol{e}_{cl}\left(\boldsymbol{T}_{cw}, \pi_{w} \right)=q\left({\pi}_{c}\right) - q\left(\boldsymbol{T}_{cw}^{-\top}{\pi}_{w}\right)
\eqno(7)
$$
Here $\pi_{w}$ is the plane parameter in the world coordinate system and $\pi_{c}$ is the observation of planes in the camera coordinate system. The camera-plane error defines the differences between the plane landmark and its corresponding observation in the camera coordinate system. The computation of the Jacobian matrix can be found in our previous work \cite{Zhang2019}.

\subsection{Data Association}
We try to match every observation of points and planes with the landmarks in the map. This process is defined as data association. Robust data association is also necessary for accurate estimation results. Point features are commonly matched by their descriptors.

To match plane features, previous works \cite{kaess2015simultaneous, CPASLAM} depend on the plane parameters $\boldsymbol{n}^\top$ and $d_\pi$ directly. This method is simple and works well for small scenes. But it needs accurate estimation of camera poses, and $d_{\pi}$ may fluctuate largely because of errors. To refine the data association of planes, we use the distances of endpoints instead.

After computing planes from intersecting lines, we save the endpoints of the lines. To match plane features, we compute the average distance $\bar{d}_{ep}$ from the endpoints to plane landmarks. Unlike $d_{\pi}$, the fluctuation of $\bar{d}_{ep}$ is relatively small. If $\bar{d}_{ep}$ is smaller than the threshold (6 cm in our experiments) and the angle between plane normal vectors is also smaller than the threshold ($12^{\circ}$ in our experiments), the plane landmark is matched with the corresponding plane observation. 

\subsection{Plane Validation}
\label{validation}
We use the same method to select keyframes as in ORB-SLAM \cite{Mur-Artal2017}. From the keyframe, every unmatched plane feature is used to create a new plane landmark. But these plane landmarks may be inaccurate, and therefore they are labelled invalid. The planes are changed to be valid only if they are observed in sufficient frames (3 keyframes in our experiments). Only valid plane features are used to estimate camera poses and added into the optimization framework (Eq. (4)).

\section{EVALUATION}
\label{evaluationSec}
In this section, we evaluate our proposed SLAM system in two popular public datasets: the EuRoC dataset \cite{EuRoC} and the KITTI vision benchmark \cite{KITTI}. The two datasets both provide stereo images. All experiments run on a laptop computer with i7-7700HQ 2.80 GHz CPU, 16GB RAM, without GPU.

We compare our proposed system with other state-of-the-art stereo SLAM systems. ORB-SLAM2 \cite{Mur-Artal2017} is a popular point-based visual SLAM system and it has a stereo camera implementation. We also compare our system with a line-based SLAM system, which utilizes the line segments directly. \cite{Qian2020} provides a SLAM system based on point and line features, and it is also built on ORB-SLAM2 and achieves state-of-the-art performance. The two SLAM systems both provide open-source codes on the Internet. Our work mainly focuses on the front-end, thus we switch off loop closure to compare drifts in our experiments.

For implementation, our system augments the stereo variant of ORB-SLAM2. We rely on the underlying ORB-SLAM2 for the point extraction and matching, and the methods to maintain a local map. The focus of our implementation is on the line segment extraction, plane computation, plane matching, and camera pose estimation using both point and plane features. In the end, we construct a global map consisting of both points and planes. To help understand the details of our system, we release our codes at https://github.com/fishmarch/Stereo-Plane-SLAM. 

\subsection{EuRoC Dataset}

The EuRoC dataset contains stereo sequences recorded from a micro aerial vehicle flying around in three different indoor environments, including an industrial machine hall and two Vicon rooms. These scenes contain man-made objects and structures, which are easy to extract line segments and compute plane features. The sequences are classified as easy, medium, and difficult according to the speed, illumination, and scene texture. The dataset also provides ground truth from the laser tracking system and the motion capture system. 

\begin{table}[t]
	\caption{Comparison of translation RMSE ($m$) in EuRoC dataset}
	\centering
	\begin{tabular}{l c c c}
		\hline
		\textbf{Sequence} & ORB-SLAM2 & Line-SLAM & Our System\\
		\hline
		MH\_01\_easy    	  		& $0.038785$			& $0.038370$		     & $\bf{0.034193}$		\\
		MH\_02\_easy		      	& $0.052046$			& $0.049897$		     & $\bf{0.048448}$		\\
		MH\_03\_medium      		& $0.037984$			& $0.048172$		     & $\bf{0.034936}$		\\
		MH\_04\_difficult		    & $0.105441$			& $\bf{0.049670}$		 & $0.067824$			\\
		MH\_05\_difficult		    & $0.092429$			& $0.055767$		     & $\bf{0.043746}$		\\
		V1\_01\_easy	            & $0.087496$			& $0.087920$		     & $\bf{0.086734}$	    \\
		V1\_02\_medium   	        & $0.091192$			& $\bf{0.064006}$		 & $0.067565$			\\
		V1\_03\_difficult		    & $0.173660$			& $\bf{0.135513}$		 & $0.165173$			\\
		V2\_01\_easy	            & $0.070456$			& $\bf{0.061183}$		 & $0.063001$			\\
		V2\_02\_medium              & $0.099885$			& $\bf{0.060296}$		 & $0.081215$			\\
		V2\_03\_difficult           & lost			        & lost		             & lost			        \\
		\hline
	\end{tabular}
	\label{EuRoc}
\end{table}

\begin{figure}[b]
	\centering
	\includegraphics[width = 7.5cm]{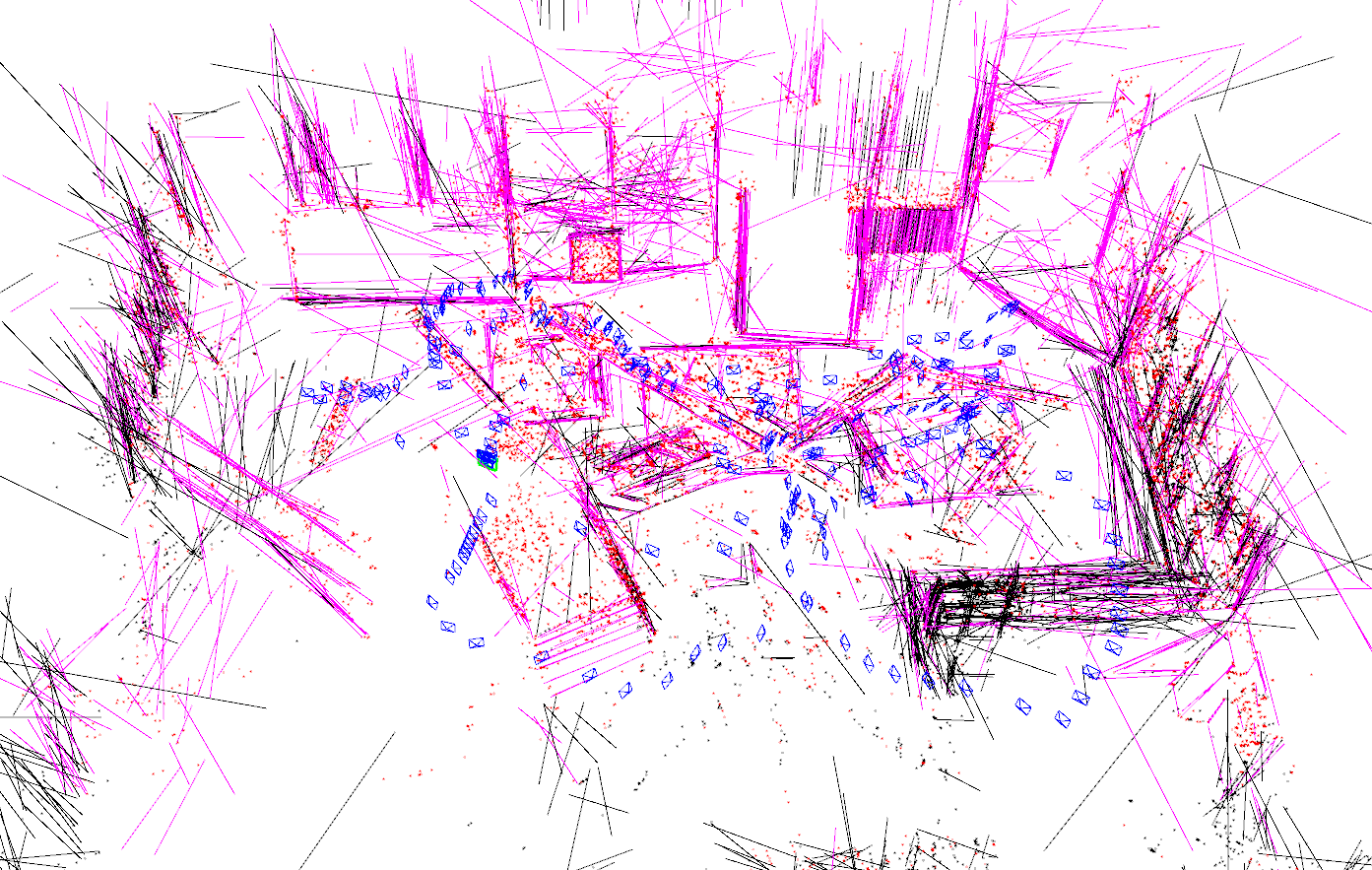}
	\caption{Built map of line-based SLAM in Sequence V1\_01\_easy.}
	\label{linemap}
\end{figure}

Table \ref{EuRoc} shows the comparison of estimation results of different SLAM systems. Here we use the absolute translation root mean square error (RMSE) to evaluate the estimation results. The smallest error for each sequence is labelled as a bold number. It is clear that our system outperforms the stereo ORB-SLAM2 in these sequences. The computed plane features bring more constraints to estimate camera poses, and these constrains are reasonable in such indoor environments.

The line-based SLAM system also performs better than ORB-SLAM2 in these sequences, and gets comparable results with ours. Although using line segments directly adds more information to the SLAM system, some line segments are not accurate and the data association is difficult when the camera view changes, as shown in Fig. \ref{linemap}. A single line in the scene may be constructed as several segments because of the failure of data association. By computing planes from line segments, our system may lose some information but filters out inaccurate line segments, and the data association of planes is easier between frames. The estimation results of our system are better in industrial machine hall sequences(MH*). We find the objects and structure in the industrial machine hall have clear and sharp edges, which benefits the line extraction and plane calculation. But in the Vicon room (V1* and V2*), there are more errors of the lines extracted from the soft cushions and curtains.

\begin{figure}[b]
	\centering
	\includegraphics[width = 8.2cm]{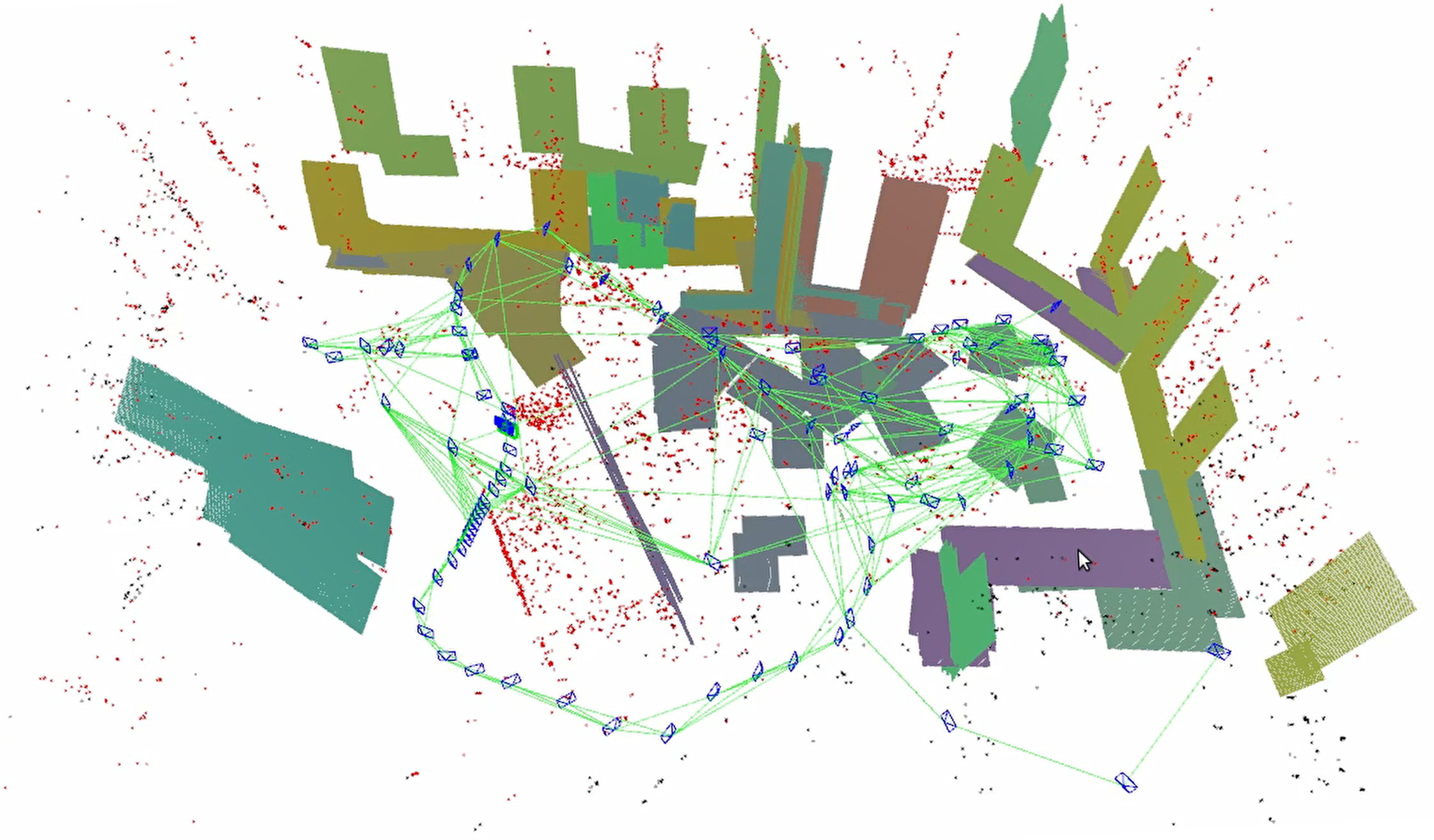}
	\caption{Built map of our system in Sequence V1\_01\_easy.}
	\label{eurocmap}
\end{figure}

An example of the built map from our system is shown in Fig. \ref{eurocmap}. The map consists of points, planes and camera poses. The plane features enrich the map and help to reflect the real scene structures clearly. It demonstrates the plane features are computed from the main objects and structure, such as walls, cushions and etc. The planes from the wall and the ground are easily matched even in frames of a large distance. Although we try to compute accurate planes and filter out those with large errors, some inaccurate planes still exist in the map. 

\begin{table}[t]
	\caption{Comparison of translation RMSE ($m$) on KITTI dataset}
	\centering
	\begin{tabular}{c c c c}
		\hline
		\textbf{Sequence} & ORB-SLAM2 & Line-SLAM & Our Method\\
		\hline
		00    	  	& $9.243199$			& $8.341743$		     & $\bf{7.929075}$		\\
		01		    & $23.313985$			& $66.529011$		     & $\bf{18.228348}$		\\
		02      	& $18.581184$			& $20.791099$		     & $\bf{18.036698}$		\\
		03		    & $9.191448$			& $9.148313$		     & $\bf{9.126736}$		\\
		04		    & $2.542802$			& $2.401305$		     & $\bf{2.207116}$		\\
		05	        & $4.756073$			& $\bf{4.403798}$		 & ${4.531347}$			\\
		06   	    & $4.959506$			& $\bf{3.710958}$		 & $4.014013$			\\
		07		    & $2.072465$			& $2.336099$		     & $\bf{1.914680}$		\\
		08	        & $15.780692$			& $\bf{13.583784}$		 & $13.705721$			\\
		09          & $7.594386$			& $\bf{7.434085}$		 & $7.642378$			\\
		10          & $6.484388$			& $\bf{6.090371}$		 & $6.19148$			\\
		\hline
	\end{tabular}
	\label{KITTI}	
\end{table}

\subsection{KITTI Dataset}
The KITTI dataset contains stereo sequences recorded from an autonomous driving platform driving in urban and highway environments. The ground truth is given by the GPS/IMU localization unit. Although our SLAM system is more suitable for indoor environments, the planes can also be computed from man-made structures (such as houses and roads) in these sequences.      

\begin{figure}[t]
	\centering
	\includegraphics[width = 6cm]{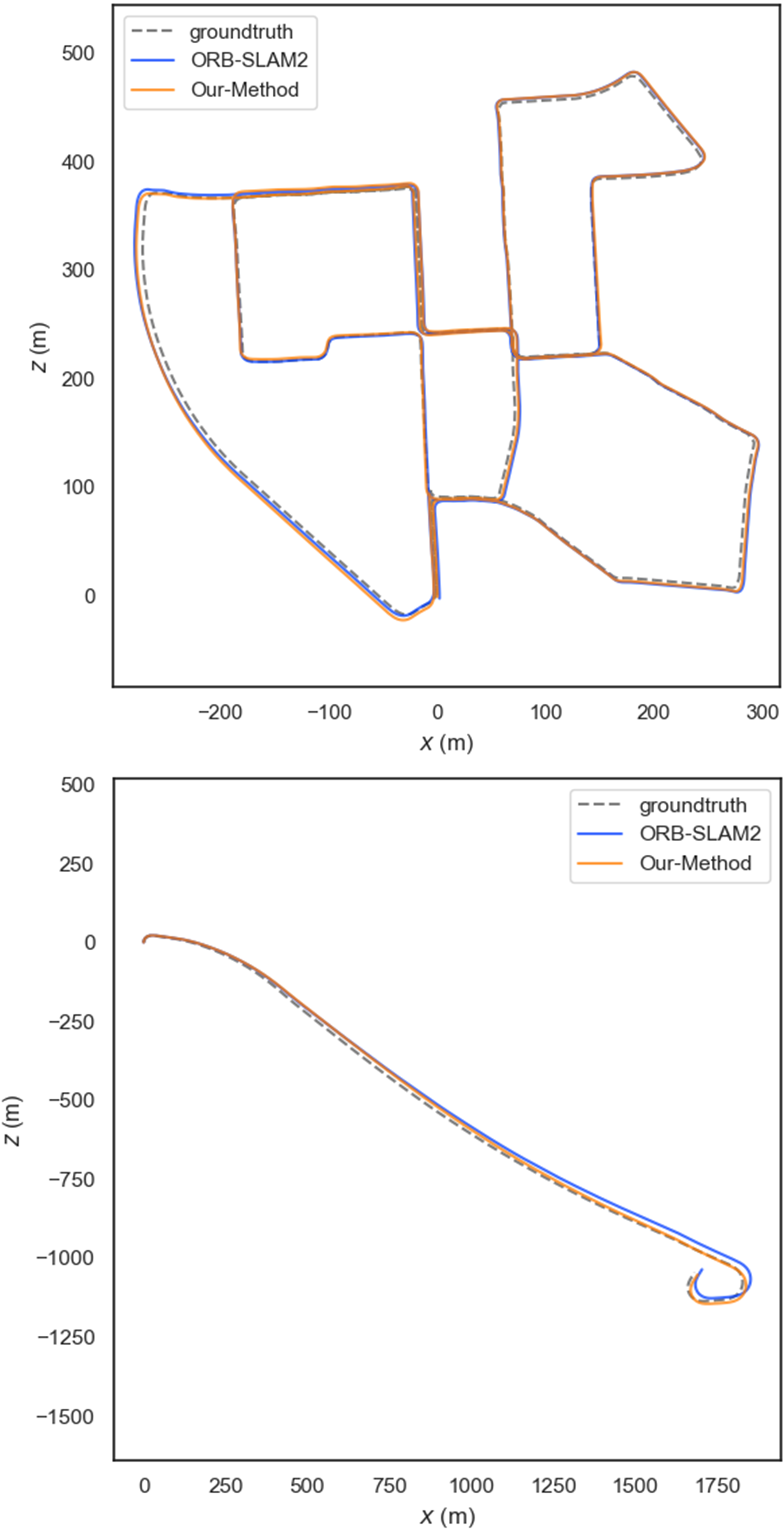}
	\caption{The comparison of trajectories in Sequence 00 and 01.}
	\label{kittiTraj}
\end{figure}

Table \ref{KITTI} shows the comparison of estimation results in the KITTI dataset. We again calculate the RMSE of these sequences and label the smallest errors as bold numbers. Our system also gets better estimation results in most of the sequences. The performance of ORB-SLAM using only point features is good enough, and our system only gets similar results in some sequences. This is because the camera moves much faster in these sequences, and thus planes are only matched in a few frames and fewer valid planes are created. Fig. \ref{kittiTraj} shows the examples of the estimated trajectories compared with the ground truth and our results are closer to the ground truth.

The line-based SLAM system also performs better in some sequences. But in other sequences, it gets even worse results, such as the Sequence 01, 02, and 07. This is because the extracted lines have large errors and are difficult to match when the camera moves quickly. 

\begin{figure}[t]
	\centering
	\includegraphics[width = 8cm]{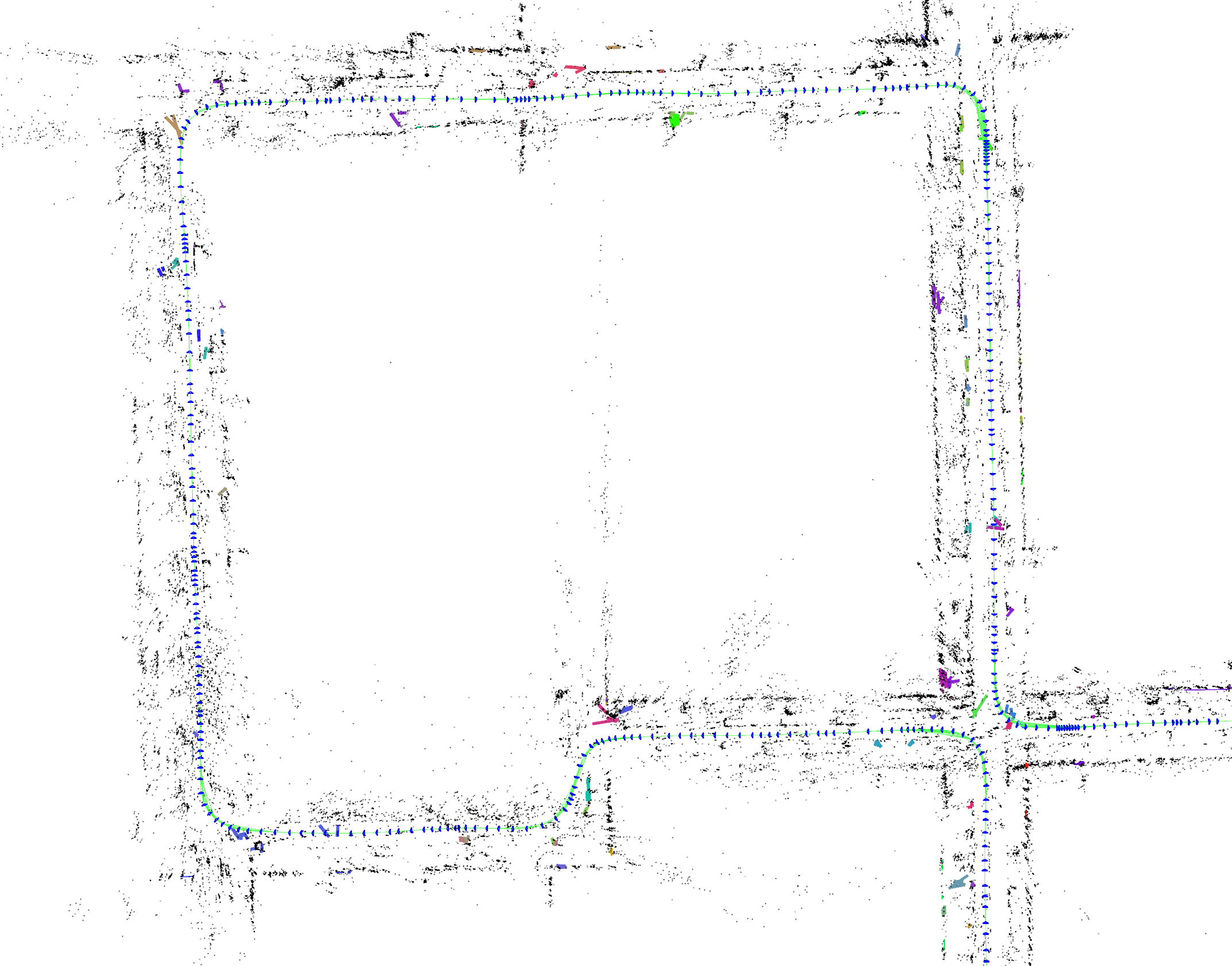}
	\caption{A part of built map in Sequence 00.}
	\label{kittimap}
\end{figure}

A part of the built map in the Sequence 00 is shown in Fig. \ref{kittimap}. The plane features are computed from roads and walls. It is clear the drift error is small. 

\subsection{Time Performance}
\begin{table}[b]
	\caption{Average processing time ($ms$) of the main parts.}
	\centering
	\begin{tabular}{c c c} 
		\hline
		& EuRoC & KITTI  \\
		{Main Part} & $752 \times 480$ & $1241 \times 376$  \\
		& 20 fps & 10 fps  \\
		\hline
		Feature Processing 			& 40.3224					& 68.9912				\\
		Feature Tracking    		& 12.5305					& 16.3811				\\
		Local  Optimization       	& 88.7514					& 110.826				\\
		\hline
		Frame Tracking 				& 54.3525					& 86.5212				\\
		\hline
	\end{tabular}
	\label{time}
\end{table}

We record the average processing time of the main parts of the proposed system, as shown in Table \ref{time}. In the table, feature processing includes point extraction, line extraction, stereo matching, and plane computation; feature tracking includes landmark matching and camera pose estimation. Frame tracking is the whole process for tracking a new coming stereo frame. The most time-consuming part is the feature processing, specifically point and line extraction. Because the images from KITTI sequences are larger, feature processing needs more time. 

We also record the average processing time of the line-based SLAM system \cite{Qian2020}. The average time for its frame tracking are $81.8247 ms$ in the EuRoC sequences and $105.347 ms$ in the KITTI sequences. It is clear that our system is faster than the SLAM system using line segments directly. In the line-based system, many line segments are built with large errors, and they are hard to be matched correctly, as shown in Fig. \ref{linemap}. But all of these features are added into the optimization function, and therefore it needs more time to estimate the optimal results.  

\section{CONCLUSIONS}

We have presented a novel method to compute plane features from stereo images for visual SLAM. Many previous works \cite{kaess2015simultaneous,CPASLAM,Zhang2019} have demonstrated the benefits of adding plane features in SLAM systems, but most of them are for RGB-D cameras. In this paper, we compute planes from stereo images instead, based on the truth that two intersecting lines determine a plane. After further validation, we add computed planes into our stereo SLAM system. We have presented our experimental results in two famous public datasets and demonstrated the accuracy of our system.

From the experimental results, it is clear that our system outperforms the state-of-the-art point-based SLAM system. Compared with the line-based SLAM system, our system also gets comparable results. The plane computing filters out inaccurate line segments and adds stable constraints to estimate camera poses. The constructed map consists of both points and planes, which reflect the real scene structures. The structured environments with man-made objects are ideal working scenes for our system.

From the constructed maps, we notice some inaccurate plane features still exist and they bring a big challenge to the data association. In the future, we would like to refine the plane computing and validation methods to get more accurate and robust plane features. In addition, we also need a more robust data association algorithm, removing the influence of estimation errors. We will also add loop closure into our system.


\end{document}